\documentclass[10pt,twocolumn,letterpaper]{article}

\usepackage{cvpr}              

\usepackage{subcaption}
\usepackage{graphicx}
\usepackage{amsmath}
\usepackage{amssymb}
\usepackage{bm}
\usepackage{booktabs,multirow}
\usepackage[accsupp]{axessibility}  
\usepackage{engord}
\usepackage{seqsplit}
\usepackage[dvipsnames]{xcolor}

\definecolor{cvprblue}{rgb}{0.21,0.49,0.74}
\usepackage[pagebackref,breaklinks,colorlinks,citecolor=cvprblue]{hyperref}

\usepackage[capitalize]{cleveref}
\crefname{section}{Sec.}{Secs.}
\Crefname{section}{Section}{Sections}
\Crefname{table}{Table}{Tables}
\crefname{table}{Tab.}{Tabs.}

\def\confName{CVPR}
\def\confYear{2024}

\begin{document}

\title{NTIRE 2025 XGC Quality Assessment Challenge: Methods and Results}

               


\author{
 Xiaohong Liu$^{*}$ \and Xiongkuo Min$^{*}$ \and Qiang Hu$^{*}$ \and Xiaoyun Zhang$^{*}$ \and  Jie Guo$^{*}$ \and Guangtao Zhai$^{*}$ \and Shushi Wang$^{*}$ \and Yingjie Zhou$^{*}$ \and   Lu Liu$^{*}$ \and Jingxin Li$^{*}$ \and Liu Yang$^{*}$ \and Farong Wen$^{*}$ \and Li Xu$^{*}$ \and Yanwei Jiang$^{*}$ \and Xilei Zhu$^{*}$ \and Chunyi Li$^{*}$ \and Zicheng Zhang$^{*}$ \and Huiyu Duan$^{*}$  \and Xiele Wu$^{*}$ \and  Yixuan Gao$^{*}$ \and 
 Yuqin Cao$^{*}$ \and Jun Jia$^{*}$ \and Wei Sun$^{*}$ \and Jiezhang Cao$^{*}$ \and Radu Timofte
 \thanks{The organizers of the NTIRE 2025 XGC Quality Challenge.\\
}
\and Baojun Li \and Jiamian Huang \and Dan Luo \and Tao Liu
\and Weixia Zhang \and  Bingkun Zheng \and Junlin Chen
\and Ruikai Zhou \and Meiya Chen \and  Yu Wang \and Hao Jiang
\and Xiantao Li \and  Yuxiang Jiang \and Jun Tang
\and Yimeng Zhao \and Bo Hu
\and Zelu Qi \and Chaoyang Zhang \and Fei Zhao \and Ping Shi
\and Lingzhi Fu \and Heng Cong \and Shuai He \and Rongyu Zhang \and Jiarong He
\and Zongyao Hu
\and Wei Luo \and  Zihao Yu \and Fengbin Guan \and Yiting Lu \and Xin Li \and Zhibo Chen
\and Mengjing Su \and  Yi Wang \and  Tuo Chen \and  Chunxiao Li \and  Shuaiyu Zhao \and Jiaxin Wen  \and  Chuyi Lin \and  Sitong Liu \and  Ningxin Chu \and Jing Wan \and  Yu Zhou
\and Baoying Chen \and  Jishen Zeng \and  Jiarui Liu \and  Xianjin Liu 
\and Xin Chen \and  Lanzhi Zhou \and  Hangyu Li \and You Han \and Bibo Xiang
\and Zhenjie Liu \and  Jianzhang Lu \and  Jialin Gui \and  Renjie Lu \and  Shangfei Wang 
\and Donghao Zhou \and  Jingyu Lin \and  Quanjian Song \and  Jiancheng Huang \and  Yufeng Yang \and  Changwei Wang 
\and Shupeng Zhong \and  Yang Yang
\and Lihuo He \and  Jia Liu \and  Yuting Xing \and  Tida Fang \and  Yuchun Jin 
}
\maketitle

\begin{abstract}
This paper reports on the NTIRE 2025 XGC Quality Assessment Challenge, which will be held in conjunction with the New Trends in Image Restoration and Enhancement Workshop (NTIRE) at CVPR 2025. This challenge is to address a major challenge in the field of video and talking head processing. The challenge is divided into three tracks, including user generated video, AI generated video and talking head.

The user-generated video track uses the FineVD-GC, which contains 6,284 user generated videos. The user-generated video track has a total of 125 registered participants. A total of 242 submissions are received in the development phase, and 136 submissions are received in the test phase. Finally, 5 participating teams submitted their models and fact sheets.

The AI generated video track uses the Q-Eval-Video, which contains 34,029 AI-Generated Videos (AIGVs) generated by 11 popular Text-to-Video (T2V) models. A total of 133 participants have registered in this track. A total of 396 submissions are received in the development phase, and 226 submissions are received in the test phase. Finally, 6 participating teams submitted their models and fact sheets. 

The talking head track uses the THQA-NTIRE, which contains 12,247 2D and 3D talking heads. A total of 89 participants have registered in this track. A total of 225 submissions are received in the development phase, and 118 submissions are received in the test phase. Finally, 8 participating teams submitted their models and fact sheets. 

Each participating team in every track has proposed a method that outperforms the baseline, which has contributed to the development of fields in three tracks.
\end{abstract}

\section{Introduction}
\label{sec:intro}
With the rapid development of video generation technologies, User-Generated Videos (UGVs), AI-Generated Videos (AIGVs), and Talking Head have become widely used in various applications. However, the quality of these videos can vary significantly due to differences in capture conditions, generation models, and animation techniques. Therefore, it is crucial to develop effective Video Quality Assessment (VQA) methods to accurately evaluate the visual quality of UGVs, AIGVs, and Talking Head, ensuring better user experience and reliable performance in real-world scenarios.
A robust quality assessment framework can help identify distortions, enhance generation techniques, and optimize models for improved visual fidelity and realism.

This NTIRE 2025 XGC Quality Assessment Challenge aims to promote the development of the quality assessment methods for videos and talking heads to guide the improvement and enhancement of the video capture, compression, and processing techniques and performance of generative models. The challenge is divided into three tracks, including user generated video track, AI generated video track and talking head track. In the user generated video track, we use the FineVD-GC~\cite{AIGIQA-20K}, which contains 6,284 user generated videos. 120 subjects are invited to produce accurate Mean Opinion Scores (MOSs). The AI generated video track uses the Q-Eval-Video~\cite{Q-Eval-100K}, in which 11 popular Text-to-Video (T2V) models are used to generate 34,029 videos. And the Sample \& Scrutinize strategy was employed during this dataset annotation process to make sure the quality and accuracy of the dataset. The talking head track uses the THQA-NTIRE \cite{zhou2024thqa,zhou2024subjective}, which contains 12,247 2D and 3D talking heads.

This challenge has a total of 347 registered participants, 125 in the user generated video track, 133 in the AI generated video track and 89 in the talking head track. A total of 863 submissions were received in the development phase, while  480 prediction results were submitted during the final testing phase. Finally, 5 valid participating teams in the user generated video track, 6 valid participating teams in the AI generated video track and 9 valid participating teams in the talking head track submitted their final models and fact sheets. They have provided detailed introductions to their quality assessment methods. We provide the detailed results of the challenge in Section~\ref{Challenge Results} and Section~\ref{Challenge Methods}. We hope that this challenge can promote the development of quality assessment in video and talking head.

This challenge is one of the NTIRE 2025~\footnote{\url{https://www.cvlai.net/ntire/2025/}} Workshop associated challenges on: ambient lighting normalization~\cite{ntire2025ambient}, reflection removal in the wild~\cite{ntire2025reflection}, shadow removal~\cite{ntire2025shadow}, event-based image deblurring~\cite{ntire2025event}, image denoising~\cite{ntire2025denoising}, XGC quality assessment~\cite{ntire2025xgc}, UGC video enhancement~\cite{ntire2025ugc}, night photography rendering~\cite{ntire2025night}, image super-resolution (x4)~\cite{ntire2025srx4}, real-world face restoration~\cite{ntire2025face}, efficient super-resolution~\cite{ntire2025esr}, HR depth estimation~\cite{ntire2025hrdepth}, efficient burst HDR and restoration~\cite{ntire2025ebhdr}, cross-domain few-shot object detection~\cite{ntire2025cross}, short-form UGC video quality assessment and enhancement~\cite{ntire2025shortugc,ntire2025shortugc_data}, text to image generation model quality assessment~\cite{ntire2025text}, day and night raindrop removal for dual-focused images~\cite{ntire2025day}, video quality assessment for video conferencing~\cite{ntire2025vqe}, low light image enhancement~\cite{ntire2025lowlight}, light field super-resolution~\cite{ntire2025lightfield}, restore any image model (RAIM) in the wild~\cite{ntire2025raim}, raw restoration and super-resolution~\cite{ntire2025raw} and raw reconstruction from RGB on smartphones~\cite{ntire2025rawrgb}.


\section{Related Work}

\subsection{User Generated VQA Dataset}
Over the years, researchers have developed various video quality assessment (VQA) datasets to analyze human visual perception characteristics. Initial datasets primarily examined synthetic degradations, employing restricted original content and manually simulated degradation patterns \cite{de2010h, moorthy2012video, wang2016mcl}. With the rise of user-generated content (UGC) platforms, contemporary research has shifted toward creating VQA databases that capture genuine quality issues encountered in practical scenarios. Multiple studies \cite{ghadiyaram2017capture, duan2024finevq, hosu2017konstanz,10960681} have specifically addressed real-world quality deterioration occurring during content capture or natural viewing environments. Other comprehensive datasets \cite{zhu2024esvqa, li2020ugc, wang2021rich, zhang2023md} have incorporated both simulated and authentic distortion types to broaden research scope. While existing UGC collections predominantly source materials from conventional platforms like YouTube, emerging datasets like KVQ \cite{lu2024kvq} specifically target short-format video content. Our proposed FineVD-GC expands this landscape by encompassing diverse video formats including on-demand streaming, conventional UGC, and short-form media. Unlike existing databases providing singular quality ratings, FineVD-GC's multi-dimensional annotations enable broader practical implementations through detailed quality characterization.

\subsection{AI Generated VQA Dataset}
Compared with user generated video quality assessment datasets, the number of proposed AI generated video (AIGV) datasets is small. Chivileva \etal~\cite{chivileva2023measuring} proposes a dataset with 1,005 videos generated by 5 T2V models. 24 users are involved in the subjective study. EvalCrafter~\cite{liu2023evalcrafter} builds a dataset using 500 prompts and 5 T2V models, resulting in 2,500 videos in total. However, only 3 users are involved in the subjective study. Similarly, FETV~\cite{liu2024fetv} uses 619 prompts, 4 T2V models, and 3 users for annotation as well. VBench~\cite{huang2024vbench} has a larger scale with in total of $\sim$1,7k prompts and 4 T2V models. Continuing with the exploration of AIGV quality assessment, the T2VQA-DB~\cite{kou2024subjective} emerges as a significant addition to the landscape. The dataset has 10,000 videos generated by 9 different T2V models. 27 subjects are invited to collect the MOSs. 
In this track, we use the latest dataset, Q-Eval-Video~\cite{Q-Eval-100K}, which contains approximately 34,000 videos generated by 11 different models. Meanwhile, the Sample \& Scrutinize strategy was employed during the dataset annotation process.

\subsection{VQA Model}

The traditional VQA models are usually designed for user-generated videos or a certain attribute of videos.~\cite{gao2023vdpve, dong2023light, kou2023stablevqa,zhang2023advancing,zhang2024reduced,liu2023ntire,han2025Interpolation, zhang2024PAPS-OVQA, kou2023stable-vqa}. For example, 
SimpleVQA \cite{sun2022deep} trains an end-to-end spatial feature extraction network to directly learn quality-aware spatial features from video frames, and extracts motion features to measure temporally related distortions at the same time to predict video quality. FAST-VQA~\cite{wu2022fast} proposes the ``fragments'' sampling strategies and the Fragment Attention Network (FANet) to accommodate fragments as inputs. Light-VQA~\cite{zhou2024Light-VQA+} and Light-VQA+~\cite{dong2023Light-VQA} provide methods for assessing the quality of videos enhanced in low-light conditions. DOVER~\cite{wu2023dover} evaluates the quality of videos from the technical and aesthetic perspectives respectively. Q-Align~\cite{wu2023qalign} can also address the VQA task by relying on the ability of multi-modal large models~\cite{zhang2024survey, zhang2024Q-Boost, wu2024Open-ended}. VQA$^2$~\cite{jia2024vqa2} further explores the approach of utilizing multi-modal large models for video quality assessment through visual question answering.

There are several works targeting the VQA tasks of AIGVs. VBench~\cite{huang2024vbench}, EvalCrafter~\cite{liu2023evalcrafter} and Q-Bench-Video~\cite{zhang2024Q-Bench-Video} build benchmarks for AIGVs by designing multi-dimensional metrics. MaxVQA~\cite{wu2023maxvqa} and FETV~\cite{liu2024fetv} propose separate metrics for the assessment of video-text alignment and video fidelity, while T2VQA~\cite{kou2024subjective} handles the features from the two dimensions as a whole. The newly emerged model, Q-Eval-Score~\cite{Q-Eval-100K} explores the use of Multimodal Large Language Models (MLLMs) for assessing the quality of AIGV. For the VQA tasks of AIGV, further research is still needed. We believe the development of these models for AIGV will certainly benefit the generation of high-quality videos.

\subsection{Talking Head}
Talking Heads is an emerging form of human-centered media, distinguished by the integration of realistic facial and vocal features \cite{zhou2023implementation,guo2024efficient}. The conventional approach to designing Talking Heads predominantly relies on facial capture technology, wherein designers utilize 3D software to map facial bones and fine-tune facial details for a specific digital persona, based on the captured facial data \cite{moura2007human,zajkac2020using}. While this manual technique can yield high-quality Talking Heads, the substantial costs associated with the required equipment, coupled with the complexity of the operation, significantly constrain the efficiency of the design process.

To address the challenges associated with Talking Head design, a range of AI-based methods have been developed. These methods can be categorized based on the type of data used to generate Talking Heads, distinguishing between generative 2D \cite{wu2018reenactgan,zakharov2019few,ha2020marionette,wiles2018x2face,wang2021safa} and generative 3D Talking Heads \cite{hong2022depth,doukas2021headgan,yao2020mesh,wang2021hififace,zhuang2022controllable,gao2023high}. Furthermore, generation techniques can be classified into vision-driven \cite{tripathy2021facegan,ren2021pirenderer,bounareli2022finding,bounareli2023stylemask,agarwal2023audio} and speech-driven \cite{sadtalker,wav2lip,audio2head,dreamtalk,iplap,videoretalking,dinet,makelttalk} approaches, depending on the fundamental principles behind their generation. Given current prominence of Talking Head generation as an active area of research, it is anticipated that more effective methods will continue to emerge.

However, existing quality assessments for Talking Heads are often limited to subjective evaluations and traditional objective quality metrics, such as PSNR and SSIM \cite{wang2004image}. While these approaches provide some insights into the quality of Talking Heads, they have notable limitations. Subjective assessments are typically time-consuming and not conducive to large-scale quantitative analysis, while objective metrics like PSNR and SSIM \cite{wang2004image} fail to capture human visual experiences and are inadequate for evaluating generative Talking Heads due to the absence of reference data. Therefore, the development of a more accurate and reliable objective quality assessment framework for Talking Heads is crucial to advancing the field of Talking Head generation.

\subsection{Digital Human Quality Assessment}
 
With the rapid advancement of digital human technology, the quality of digital humans has garnered significant attention. To explore this issue in greater depth, Zhang $et$ $al.$ have developed several datasets, including DHHQA \cite{zhang2023perceptual}, DDH-QA \cite{zhang2023ddh}, and SJTU-H3D \cite{zhang2023advancing}, focusing on captured 3D digital humans. These datasets provide rich data for assessing the quality of static heads, dynamic full-body digital humans , and static full-body digital humans. Additionally, they have designed full-reference \cite{zhang2023perceptual}, reduced-reference \cite{zhang2024reduced}, and no-reference evaluation methods for these datasets, incorporating siamese networks \cite{zhang2023perceptual}, multi-task learning \cite{zhou2023no}, and multi-modal information fusion \cite{chen2023no,zhang2023geometry} techniques. These approaches not only offer reliable assessment frameworks for various types of digital humans but also account for different applicability scenarios. Furthermore, to investigate potential quality degradation during communication transmission, Zhou $et$ $al.$ and Zhang $et$ $al.$ have conducted user experience quality assessments for 3D talking heads and 3D talking digital humans, respectively. They first established the THQA-3D \cite{zhou2024subjective} and 6G-DTQA \cite{zhang2024quality} datasets and proposed corresponding objective evaluation algorithms that integrate channel parameters, visual features, and audio features. Despite these advancements, existing datasets for digital human quality assessment are often constrained by limited data size and insufficient diversity of digital human models, which in turn restricts the generalizability of assessment algorithms.

In recent years, the rapid growth of generative AI has enabled more efficient solutions for designing and acquiring digital humans \cite{zhou2024reli,zhou20253dgcqa,zhou2024memo}. In response, Zhou $et$ $al.$ developed the first THQA dataset \cite{zhou2024thqa} for speech-driven Talking Heads. This dataset includes 800 Talking Heads generated by applying eight representative speech-driven algorithms to 20 images. While this dataset introduces the Talking Head Quality Assessment challenge, it unfortunately does not provide a reliable quality assessment framework. To address this gap, the present work seeks to establish a comprehensive evaluation scheme for this emerging media by engaging experts in discussions on the development of an appropriate assessment methodology.

\section{NTIRE 2025 XGC Quality Assessment Challenge}
We organize the NTIRE 2025 XGC Quality Assessment Challenge, including user generated video quality assessment, AI generated video quality assessment and talking head quality assessment, in order to promote the development of objective quality assessment methods. The main goal of the challenge is to predict the perceptual quality of videos and talking heads. Details about the challenge are as follows:

\subsection{Overview}
The challenge has three tracks, \ie user generated video track, AI generated video track and talking head track. The task is to predict the perceptual quality of video and talking head based on a set of prior examples and their perceptual quality labels. The challenge uses FineVD-GC~\cite{duan2024finevq} , the Q-Eval-Video~\cite{Q-Eval-100K} and THQA\cite{zhou2024thqa,zhou2024subjective} dataset and splits them into the training, validation, and testing sets. As the final result, the participants in the challenge are asked to submit predicted scores for the given testing set.

\subsection{Datasets}

\begin{table*}[!t]
\Large
    \centering
    \caption{Quantitative results for the NTIRE 2025 XGC Quality Assessment Challenge: Track 1 User Generated Video. \textit{Color}, \textit{Noise}, \textit{Artifact}, \textit{Blur}, \textit{Temporal}, and \textit{Overall} indicate the main scores for each dimension.}
    \resizebox{\textwidth}{!}{
    \begin{tabular}{c|c|c|cccccc|ccc}
    \toprule
    Rank & Team & Leader &       Color&Noise&Artifact&Blur&Temporal&Overall&Main Score & SRCC & PLCC \\
    \midrule
    1 & SLCV & Baojun Li &       0.8898&0.8411&0.8805&0.9101&0.8216&0.8954&0.8731& 0.8724& 0.8738\\
    2 & SJTU-MOE-AI &  Weixia Zhang &        0.8734&0.8327&0.8706&0.8880&0.8237&0.8836&0.8620 &  0.8591&  0.8649\\
    3 & MiVQA & Ruikai Zhou &       0.8655&0.8136&0.8467&0.8695&0.8055&0.8636&0.8440& 0.8386& 0.8494\\
    4 & XGC-Go & Xiantao Li &       0.8512&0.7906&0.8353&0.8575&0.7623&0.8521&0.8248& 0.8222& 0.8273\\
    5 & FoodVQA& Yimeng Zhao &       0.8390&0.7773&0.8239&0.8527&0.7633&0.8415&0.8162& 0.8125& 0.8199\\
    \midrule
    \multirow{1}{*}{Baseline} & \multicolumn{2}{c|}{FastVQA~\cite{wu2022fastvqaefficientendtoendvideo}} &       0.7982&			0.7476&		0.7929&0.7988&0.7325&		0.8038&0.7789& 0.7740& 0.7837\\
    \bottomrule
    \end{tabular}
    }
    \label{tab:user generated video results}
\end{table*}

\begin{table*}[!t]
    \centering
    \caption{Quantitative results for the NTIRE 2025 Quality Assessment of AI-Generated Content Challenge: Track 2 AI Generated Video.}
    \resizebox{0.8\textwidth}{!}{
    \begin{tabular}{c|c|c|ccc}
    \toprule
    Rank & Team & Leader & Main Score & SRCC & PLCC \\
    \midrule
    1 & SLCV & Baojun Li & 0.6645 & 0.6621 & 0.6669 \\
    2 & CUC-IMC & Zelu Qi & 0.6310 & 0.6080 & 0.6539 \\
    3 & opdai & Lingzhi Fu & 0.5903 & 0.5854 &  0.5952\\
    4 & Magnolia & Zongyao Hu & 0.5889 & 0.5933 & 0.5844  \\
    5 & AIGC VQA & Wei Luo & 0.5606 & 0.5485 & 0.5727 \\
    6 & SJTU-MOE-AI & Bingkun Zheng & 0.5463 & 0.5530 & 0.5396 \\

    \midrule
    \multirow{3}{*}{Baseline} & \multicolumn{2}{c|}{Q-Eval-Score~\cite{Q-Eval-100K}} & 0.4741 & 0.4861 & 0.4642 \\
    ~ & \multicolumn{2}{c|}{DOVER~\cite{wu2023dover}} & 0.5055 & 0.5057 & 0.5054\\
    ~ & \multicolumn{2}{c|}{T2VQA~\cite{kou2024subjective}} & 0.5161 & 0.5161 & 0.5160 \\
    \bottomrule
    \end{tabular}
    }
    \label{tab:AI generated video results}
\end{table*}

\begin{table*}[!t]
    \centering
    \caption{Quantitative results for the NTIRE 2025 XGC Quality Assessment: Track 3 Talking Head. }
    \resizebox{0.8\textwidth}{!}{
    \begin{tabular}{c|c|c|ccc}
    \toprule
    Rank & Team & Leader & Main Score & SRCC & PLCC \\
    \midrule
    1& QA Team & Mengjing Su&0.8244 & 0.8036 & 0.8453 \\
    2& MediaForensics& Baoying Chen&0.8236 & 0.8024 & 0.8448 \\
    3& AutoHome AIGC & Xin Chen&0.8046 & 0.7864 & 0.8229 \\
    4& USTC-AC & Zhenjie Liu&0.8044 & 0.7813 & 0.8275 \\
    5& SJTU-MOE-AI &Junlin Chen &0.8003 & 0.7797 & 0.8209 \\ 
    6& FocusQ & Donghao Zhou & 0.7921 & 0.7708 & 0.8135 \\ 
    7& NJUST-KMG & Shupeng Zhong &0.7896 & 0.7599 & 0.8193 \\
    8& XIDIAN-VQATeam & Lihuo He&0.7872 & 0.7730 & 0.8015 \\ 
    \midrule
    \multirow{1}{*}{Baseline} & \multicolumn{2}{c|}{SimpleVQA~\cite{sun2022a}} & 0.7862 & 0.7662 & 0.8062\\
    \bottomrule
    \end{tabular}
    }
    \label{tab:talking head results}
\end{table*}

In the user-generated video track, we use a new dataset called ``Fine-grained Video Database - Generated Content'' (FineVD-GC) \cite{duan2024finevq}, which comprises a total of 6,284 web-crawled UGC videos sourced from YouTube, TikTok, and Bilibili. Each video is randomly clipped into 8-second segments. We initially employ FastVQA \cite{wu2022fast} to assess the video quality. Based on the distribution of the quality scores, we uniformly sample videos that span a wide range of categories and exhibit a diverse spectrum of quality. These videos are subsequently manually filtered to ensure a comprehensive representation of various distortion types. 120 subjects are invited to rate the videos in FineVD-GC. After normalizing and averaging the subjective opinion scores, the mean opinion score (MOS) of each video can be obtained. Furthermore, we randomly split the FineVD-GC into a training set, a validation set, and a testing set according to the ratio of $4:1:1$. The numbers of videos in the training set, validation set, and testing set are $4,190$, $1,047$, and $1,047$, respectively.

In the AI generated video track, we use the Q-Eval-Video~\cite{Q-Eval-100K}. The dataset contains 34,000 generated videos from: CogVideoX~\cite{Yang2024CogvideoX}, GEN-2~\cite{2024Gen2}, GEN-3~\cite{2024Gen3}, Latte~\cite{Ma2024Latte}, Kling~\cite{2024Kling}, Dreamina~\cite{2023Dreamina}, Luma~\cite{2024Luma}, PixVerse~\cite{2024Pixverse}, Pika~\cite{2024Pika}, SVD~\cite{Blattmann2023SVD} and Vidu~\cite{2024Vidu}. These videos were generated using approximately 4,700 prompts sampled from VBench, EvalCrafter, T2VCompench, and VideoFeedback. Every video resolution is unified to $512 \times 512$, and the video length is 2s. 

In the Talking Head track, we utilize the THQA-NTIRE dataset for training, validation, and testing. This dataset integrates and extends the existing THQA \cite{zhou2024thqa} and THQA-3D \cite{zhou2024subjective} datasets, comprising a total of 12,247 Talking Heads. Specifically, it includes 11,247 generative 2D Talking Heads and 1,000 3D Talking Heads, providing a comprehensive dataset for the development of a unified Talking Head quality assessment framework. All Talking Heads in the dataset contain audio information and exhibit a diverse range of resolutions and durations, thereby posing increased challenges for accurate and robust quality assessment.


\subsection{Evaluation Protocol}

In both tracks, the main scores are utilized to determine the rankings of participating teams. We ignore the sign and calculate the average of Spearman Rank-order Correlation Coefficient (SRCC) and Person Linear Correlation Coefficient (PLCC) as the main score:
\begin{equation}
    \mathrm{Main\;Score} = (|\mathrm{SRCC}| + |\mathrm{PLCC}|)/2.
\end{equation}
SRCC measures the prediction monotonicity, while PLCC measures the prediction accuracy. Better quality assessment methods should have larger SRCC and PLCC values. Before calculating PLCC index, we perform the third-order polynomial nonlinear regression. By combining SRCC and PLCC, the main scores can comprehensively measure the performance of participating methods.
\begin{figure*}[ht]
    \centering
    \begin{subfigure}{1\textwidth}
        \centering
        \includegraphics[width=\textwidth]{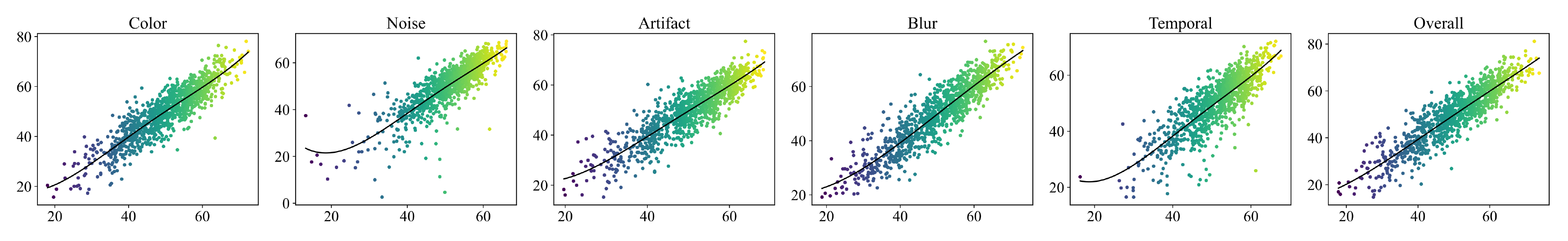}
        \caption{SLCV}  
        \label{fig:rank1}
    \end{subfigure}
    \begin{subfigure}{1\textwidth}
        \centering
        \includegraphics[width=\textwidth]{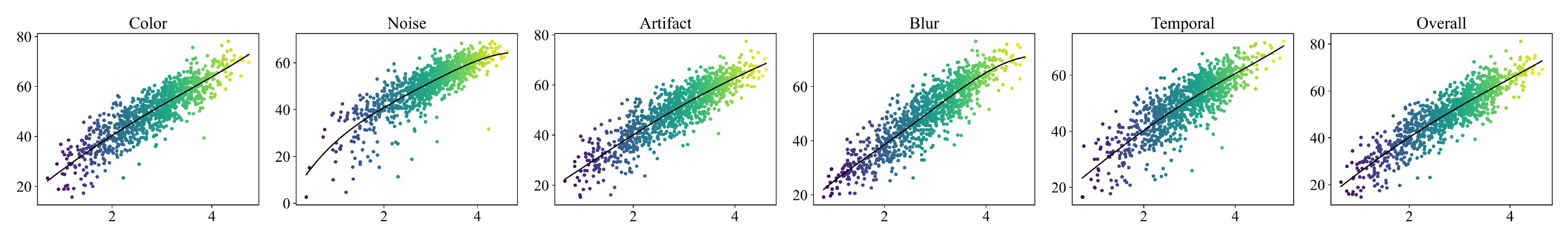}
        \caption{SJTU-MOE-AI}  
        \label{fig:rank2}
    \end{subfigure}
    \begin{subfigure}{1\textwidth}
        \centering
        \includegraphics[width=\textwidth]{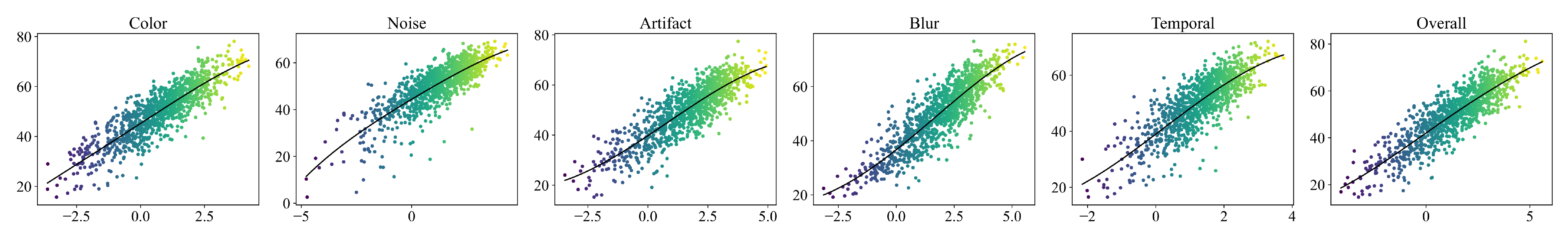}
        \caption{MiVQA}  
        \label{fig:rank3}
    \end{subfigure}
    \begin{subfigure}[b]{1\textwidth}
        \centering
        \includegraphics[width=\textwidth]{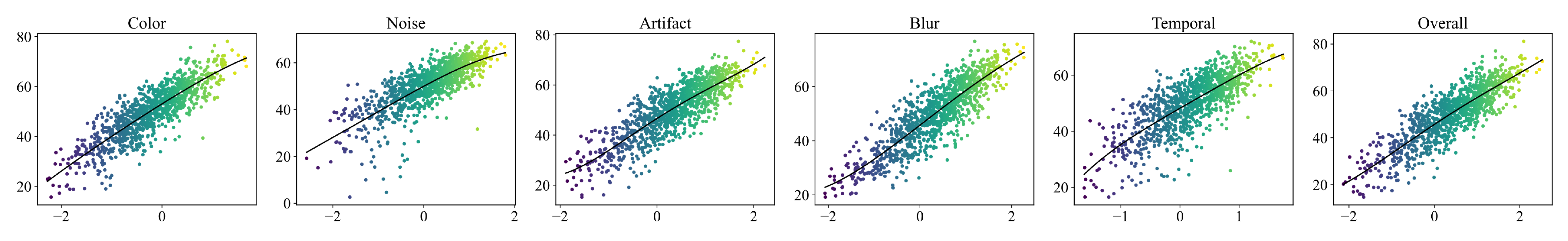}
        \caption{XGC-Go}  
        \label{fig:rank4}
    \end{subfigure}
    \begin{subfigure}[b]{1\textwidth}
        \centering
        \includegraphics[width=\textwidth]{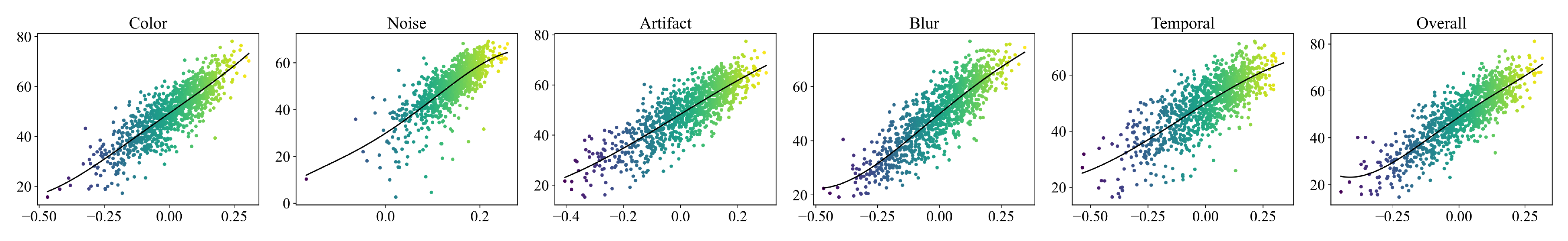}
        \caption{FoodVQA}  
        \label{fig:rank5}
    \end{subfigure}
    
    \caption{Scatter plots of the predicted scores vs. MOSs in the user-generated video track. The curves are obtained by a four-order polynomial nonlinear fitting.}
    
    \label{fig:track1-scatter}
\end{figure*}
\subsection{Challenge Phases}
Both tracks consist of two phases: the developing phase and the testing phase. In the developing phase, the participants can access the generated images/videos of the training set and the corresponding prompts and MOSs. Participants can be familiar with dataset structure and develop their methods. We also release the generated images and videos of the validation set with the corresponding prompts but without corresponding MOSs. Participants can utilize their methods to predict the quality scores of the validation set and upload the results to the server. The participants can receive immediate feedback and analyze the effectiveness of their methods on the validation set. The validation leaderboard is available. In the testing phase, the participants can access the images and videos of the testing set with the corresponding prompts but without corresponding MOSs. Participants need to upload the final predicted scores of the testing set before the challenge deadline. Each participating team needs to submit a source code/executable and a fact sheet, which is a detailed description file of the proposed method and the corresponding team information. The final results are then sent to the participants.

\begin{figure*}[ht]
\centering
\includegraphics[width=\linewidth]{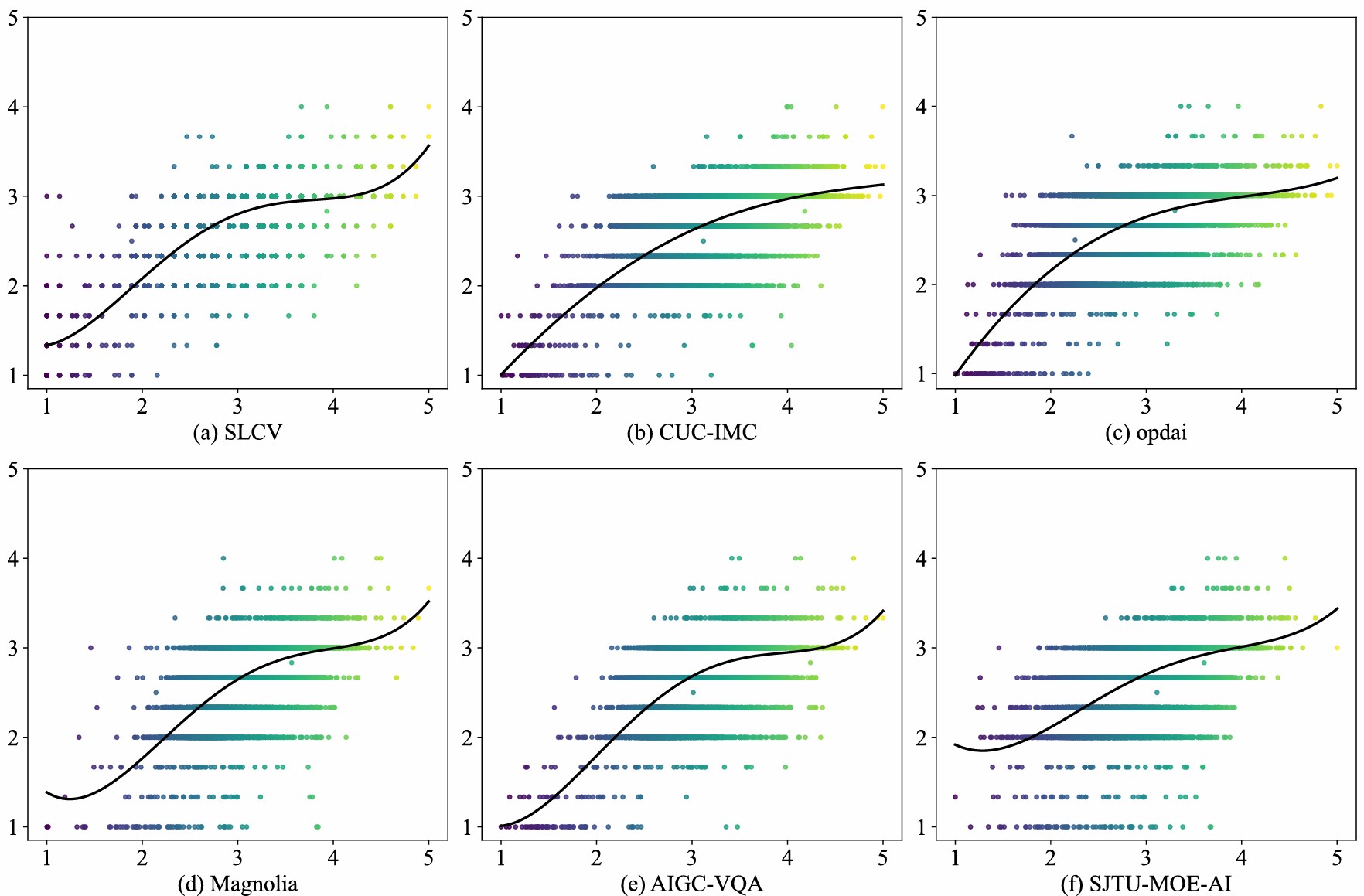}

\caption{Scatter plots of the predicted scores vs. MOSs in the AI generated video track. The curves are obtained by a four-order polynomial nonlinear fitting.}
\label{fig:track2 AI-generated video}
\end{figure*}

\begin{figure*}
    \centering
    \includegraphics[width=0.8\linewidth]{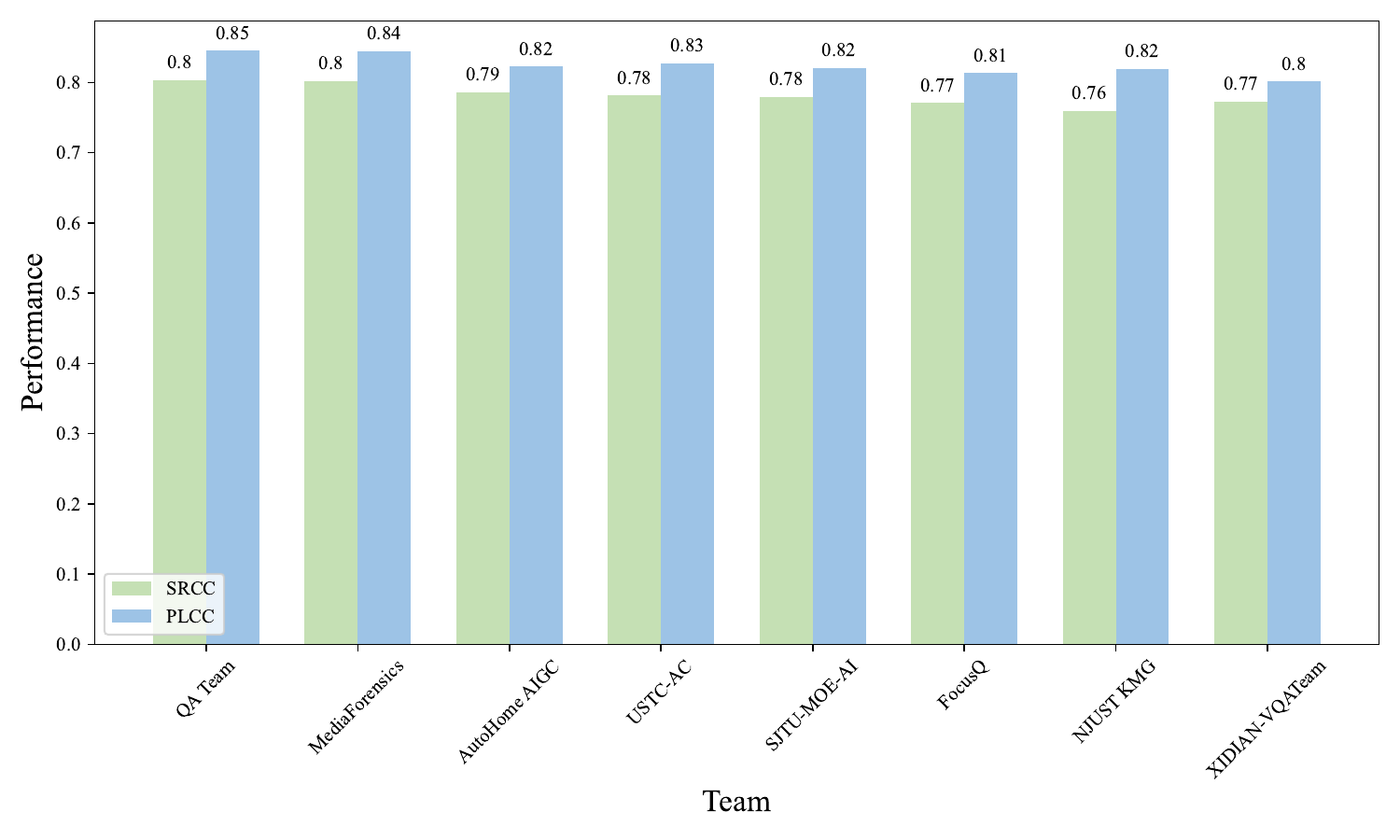}
    \caption{The performance of methods proposed by different teams in Talking head track.}
    \label{fig:track3-total}
\end{figure*}

\section{Challenge Results}
\label{Challenge Results}

5 teams in the user generated video track, 6 teams in the AI generated video track and 8 teams in the talking head track have submitted their final codes/executables and fact sheets. 
Table~\ref{tab:user generated video results}, Table~\ref{tab:AI generated video results} and Tabel ~\ref{tab:talking head results} summarize the main results and important information of the 19 valid teams. Detailed information about all participating teams and their algorithms can be found in the supplementary materials.

\subsection{Baselines}
We compare the performance of submitted methods with several quality assessment methods on the testing set, including FastVQA~\cite{wu2022fastvqaefficientendtoendvideo}, Q-Eval-Score~\cite{Q-Eval-100K}, DOVER~\cite{wu2023dover}, T2VQA~\cite{kou2024subjective} and SimpleVQA~\cite{sun2022a} for these three tracks.

\subsection{Result Analysis}
The main results of 19 teams' methods and the baseline methods are shown in Table~\ref{tab:user generated video results}, Table~\ref{tab:AI generated video results} and Table~\ref{tab:talking head results}. It can be seen that in three tracks, the results of the baseline methods are not all ideal in the testing set of three datasets, while most of the submitted methods have achieved better results. It means that these methods are closer to human visual perception when used to evaluate the content. In the user generated video track, 5 teams all achieve a main score higher than 0.8, and 2 teams are higher than 0.85. In the AI generated video track, 6 teams achieve a main score higher than 0.5, 2 teams higher than 0.6, and the championship team is higher than 0.65. In the talking head track, 7 teams achieve a main score higher than baseline, and 5 teams higher than 0.8. In the meantime, the top-ranked teams only have a small difference in the main score. Figures~\ref{fig:track1-scatter} and~\ref{fig:track2 AI-generated video} show scatter plots of predicted scores versus MOSs for the 10 teams' methods on the testing set. The curves are obtained by polynomial nonlinear fitting. We can observe that the predicted scores obtained by the top team methods have higher correlations with the MOSs. In track 3, Figure~\ref{fig:track3-total} more intuitively shows the performance of the 8 teams' methods. These results demonstrate the effectiveness of the submitted methods in improving quality assessment across all tracks, highlighting their potential for better alignment with human perception.


\section{Challenge Winner Methods}
\label{Challenge Methods}

\subsection{User-generated Video Track}

\label{pengfei}

Team SLCV wins the championship in the user-generated video track. Unlike conventional approaches that rely on regression
or classification for video quality assessment (e.g., LIQE~\cite{zhang2023liqe}, Q-Align~\cite{wu2023qalign}, Fast-VQA    ~\cite{wu2022fastvqaefficientendtoendvideo}, and SimpleVQA~\cite{sun2022a}),
their method leverages a multimodal large language model
(MLLM) to estimate video quality. In InternVL 2.5~\cite{cai2024internlm2},
an effective data filtering process was introduced, leveraging large language model (LLM) scoring to evaluate and
remove low-quality samples, thereby improving the overall
quality of the training data. Inspired by this capability of
InternVL 2.5 to assess data quality using LLM-based scoring, they adopt a multimodal large language model (MLLM)
for estimating video quality in our work. Specifically, they
directly utilize the InternVL 2.5 model as the MLLM to
achieve robust and reliable video quality assessment .
To overcome
the limitation in the spatial domain, they introduce Spatial
Window Sampling as a data augmentation strategy. Specifically, they employ a sliding window approach that crops
the original video frames with a window size set to 3/4
of the video’s longest side. This method effectively triples
the amount of training data, thereby enhancing the model’s
ability to learn fine-grained spatial features.
They employ the LoRA (Low-Rank Adaptation) method to
efficiently fine-tune the InternVL 2.5 model, enabling it to
perform the six fine-grained quality assessments. During inference,
the same data processing strategy used during training is
applied to the test videos. Specifically, the model independently predicts quality scores for the three sub-videos generated by the sliding window sampling process. The final
prediction is then obtained by averaging the results across
these sub-videos. This approach not only ensures robust
training but also facilitates accurate and reliable evaluation of fine-grained video quality.\\

\subsection{AI Generated Video Track}
Team SLCV is the final winner of the AI generated video track. They propose temporal pyramid sampling, to address the unique challenges posed by AI-generated videos in quality assessment. Unlike user generated video, the quality assessment of these AI generated videos primarily focuses on two core aspects: the smoothness of object motion and the authenticity of the content. To effectively capture these critical metrics, the team design the temporal pyramid sampling method to capture the dynamic characteristics of videos at multiple temporal resolutions. This is achieved by performing multi-scale frame interval sampling at varying frequencies. The original video is sampled at different frame rates and lengths, generating multiple subsets of data with diverse temporal granularities. Each subset is then used to independently train the model, enabling it to learn distinct motion smoothness and content authenticity features at different temporal scales.

\subsection{Talking Head Track}

\label{QA Team}
The QA Team wins the champion in the Talking Head (TH) track. They proposed a novel NR video quality assessment model based on multimodal feature representations, comprising four modules: spatial feature extraction, temporal feature extraction, audio feature extraction, and audio-visual fusion. Visual distortions are categorized into spatial and motion distortions. The types of visual distortions in videos can be roughly divided into two categories: spatial distortion\cite{zhou2023} and motion distortion. First, Talking Head videos are split into clips for spatial and temporal feature extraction.Whole clip is utilized for temporal feature extraction with a fixed pretrained 3D-CNN backbone SlowFast\cite{feichtenhofer2019slowfast}.The first frame of each clip is used for spatial feature extraction.The spatial feature extraction module utilizes an efficient channel attention module ECA-Net\cite{30ECA}, to effectively achieve cross-channel interaction,and then utilize the SwinTransformer-tiny\cite{liu2021swin} to extract visual features from the first frame.

For audio feature extraction, the audio is aligned with the visual frames, and four techniques—chromagram, CQT, MFCC, and GFCC—are used to extract time-frequency features. These features are stacked into 4 channels and fed into a separable convolution network with frequency, time, and fusion blocks, each consisting of Conv2D layers, BatchNorm, and Maxpool. The frequency and time blocks use 1×m and n×1 kernels, respectively, to perform spatially separable convolutions, reducing parameters. Temporal information is processed using Bi-LSTM, which captures context from both past and future sequences. Finally, the features are fused into a quality score using fully connected (FC) layers.

Videos are divided into 1-second clips, with 6 clips selected via cyclic sampling.  The Swin Transformer extracts spatial features with 3×224×224 patches, while SlowFast extracts temporal features from resized 224×224 clips.

\section*{Acknowledgments}

This work was partially supported by the Humboldt Foundation. We thank the NTIRE 2025 sponsors: ByteDance, Meituan, Kuaishou, and University of Wurzburg (Computer Vision Lab).


{\small
\bibliographystyle{ieee_fullname}
\bibliography{egbib}
}

\end{document}